\documentclass[10pt,twocolumn,letterpaper]{article}

\usepackage[pagenumbers]{cvpr} % To force page numbers, e.g. for an arXiv version

\usepackage{graphicx}
\usepackage{algorithm}
\usepackage{algorithmic}
\usepackage{tikz}
\usetikzlibrary{arrows.meta,calc,fit,positioning}

\usepackage{array}
\usepackage{booktabs}
\usepackage{multirow}
\newcolumntype{L}[1]{>{\raggedright\arraybackslash}p{#1}}
\usepackage{stfloats}  % allow figure* on same page as definition
\usepackage{caption}   % for \captionof in non-float environments

\definecolor{cvprblue}{rgb}{0.21,0.49,0.74}
\usepackage[breaklinks,colorlinks,allcolors=cvprblue]{hyperref}

\def\paperID{*****} % *** Enter the Paper ID here
\def\confName{CVPR}
\def\confYear{2026}

\title{GroundShot: Visually Consistent Multi-Shot Long Video Generation via Entity-Grounded Shot Scheduling}

\author{Yixuan Lai\\
State Key Lab of CAD\&CG\\
Zhejiang University\\
Hangzhou, China\\
{\tt\small yixuan.lai@zju.edu.cn}
\and
Tianjia Shao\thanks{Corresponding author.}\\
State Key Lab of CAD\&CG\\
Zhejiang University\\
Hangzhou, China\\
{\tt\small tjshao@zju.edu.cn}
\and
Weijia Dou\\
Fudan University\\
Shanghai, China\\
{\tt\small dwj132705@gmail.com}
\and
Siyu Zhu\\
Fudan University\\
Shanghai, China\\
{\tt\small siyuzhu@fudan.edu.cn}
\and
Jingdong Wang\\
Baidu\\
Beijing, China\\
{\tt\small wangjingdong@outlook.com}
}

\begin{document}
\twocolumn[{%
\maketitle
\begin{center}
    \includegraphics[width=\textwidth]{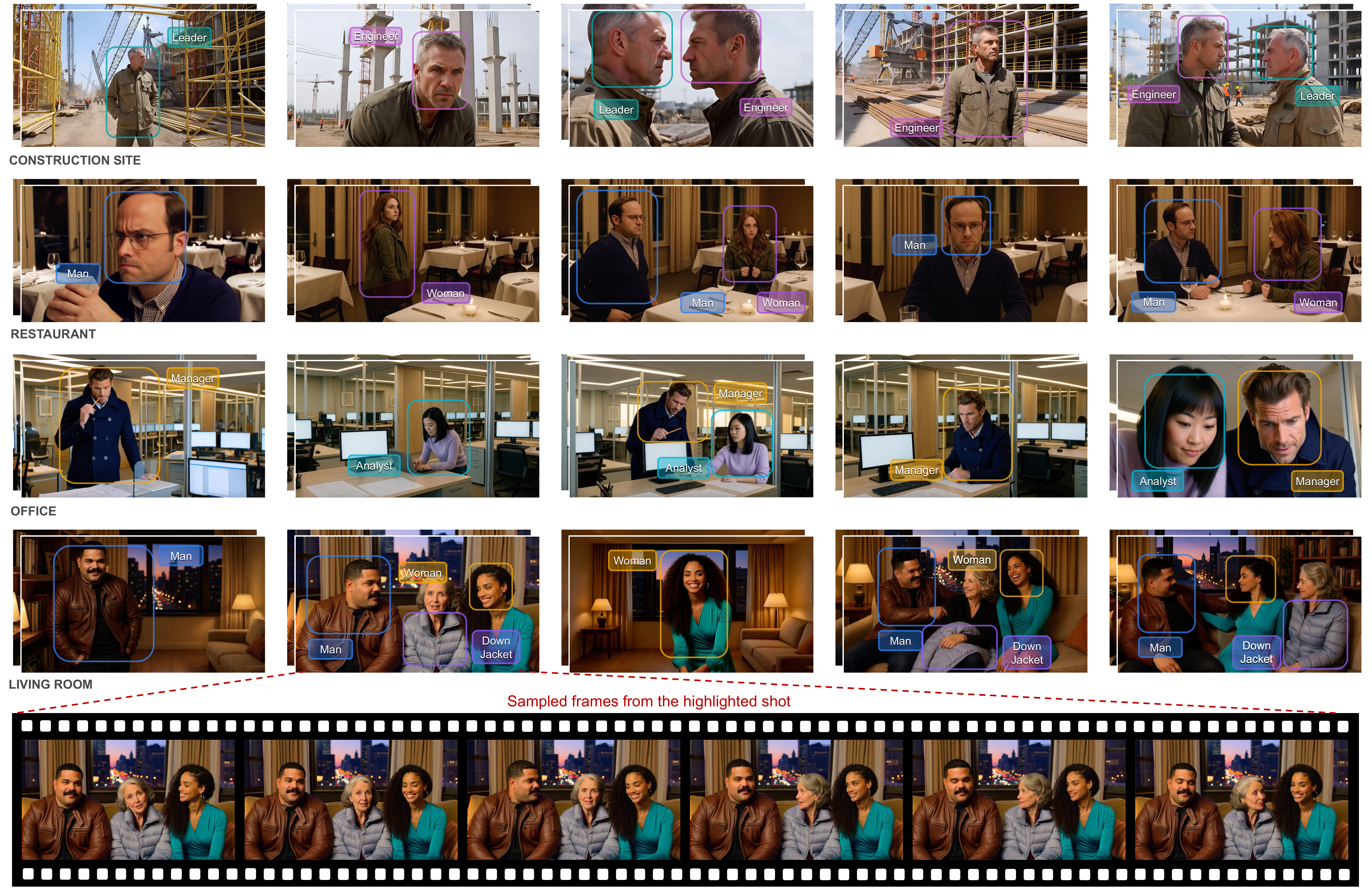}
    \captionof{figure}{\textbf{GroundShot generates visually consistent multi-shot long videos from text scripts.}
    Each row shows a generated video with colored annotations marking recurring characters and objects whose cross-shot consistency should be compared.
    Despite changes in viewpoint, scale, composition, and interaction, GroundShot preserves entity identity, clothing and object details, and scene appearance.
    The bottom strip expands one highlighted shot into sampled frames, illustrating temporal continuity and natural within-shot dynamics.}
    \label{fig:teaser}
\end{center}\vspace{1em}
}]

\begin{abstract}

Generating visually consistent multi-shot videos remains an open challenge. As videos span more shots, inconsistencies can accumulate across shots, causing entities that reappear across shots---characters, objects, and locations---to drift away from how they first appear. We observe that viewers judge consistency by comparing each later appearance of an entity with its first clear appearance; the visual quality of this initial appearance sets the consistency ceiling for all that follows.
Motivated by this, we present \textbf{GroundShot}, a training-free, model-agnostic agentic framework for entity-grounded multi-shot generation. GroundShot builds an entity-level visual memory online from accepted generated shots: it schedules shots' generation order by their expected usefulness as entity references, grounds entities from generated videos, verifies their reliability before adding them to memory, and retrieves suitable entity references from memory before each shot is generated.
To evaluate this entity-centered view of consistency, we further introduce \textbf{GroundBench}, a diagnostic benchmark that measures consistency at the entity level while isolating controlled challenge dimensions. Experiments show that GroundShot improves multi-shot consistency over existing methods while requiring no additional training or model modification.

\end{abstract}
    
% ============================================================================
\section{Introduction}

Recent video generation models can generate visually consistent short clips, but extending them to coherent multi-shot narratives remains difficult.
As the number of shots grows, visual continuity becomes harder to maintain--—characters may shift in facial features, objects may change in color or shape, and locations may lose spatial coherence.
Such drift is especially problematic for sequential consistency mechanisms, where later shots rely on preceding frames, keyframes, or frame-level memories derived from earlier outputs; errors in this historical context can propagate forward and eventually lead to substantial identity or scene drift.

Existing multi-shot methods often preserve visual continuity through nearby frames or frame-level memories~\cite{filmweaver,shotstream,moviedreamer,storymem,videomemory,a2rd}.
Yet two limitations remain.
First, whole-frame references entangle characters, objects, and locations, forcing the generator to infer each entity's appearance from mixed visual context, which becomes increasingly unreliable as scenes grow more complex.
Second, generation usually follows narrative order, so memory may be initialized from an early weak view even when a later shot would provide a clearer reference. Later shots conditioned on this low-quality memory inherit a weak visual constraint, making recurring entities difficult to preserve consistently across the shot sequence.

\begin{figure}[t]
    \centering
    \includegraphics[width=\columnwidth]{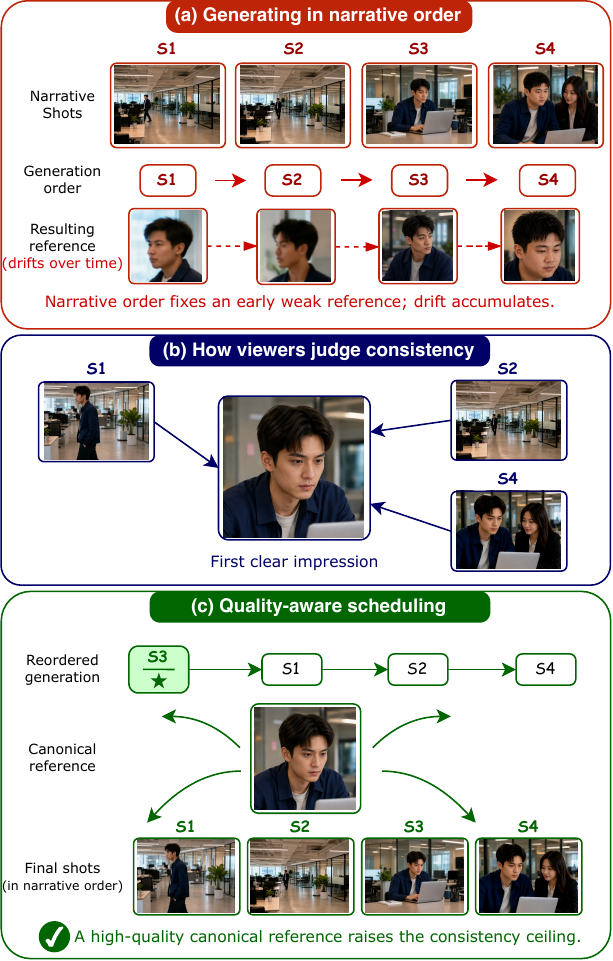}
    \caption{\textbf{Motivation for quality-aware shot scheduling.}
    \textbf{(a)} Narrative-order generation may build memory from an early low-quality entity crop, leaving later shots with a weak reference.
    \textbf{(b)} Viewers judge recurring entities against a canonical reference, motivating a star-shaped consistency structure that ties all appearances to the same reference.
    \textbf{(c)} GroundShot first generates shots likely to yield clear entity-level references, keeps only reliable grounded crops, and uses them to condition dependent shots before restoring narrative order.}
    \label{fig:motivation}
\end{figure}

These two limitations point to two requirements.
The first limitation suggests that consistency should be managed at the entity level, so each character, object, and location can be localized and referenced separately.
This requires \emph{entity-level grounding}: after a shot is generated, the system should identify the regions corresponding to each recurring entity and turn reliable ones into entity references.
The second limitation suggests that shot generation order need not follow shot narrative order. Instead, shots that are more suitable for establishing references can be generated before other shots depend on them.
This matches human perception: once viewers form their first clear impression of a character, object, or location, they remember how it looks and judge later appearances against that impression. We refer to this standard as the \emph{canonical reference}. Every occurrence of that entity, regardless of its narrative position, should remain faithful to the same canonical reference. This yields a star-shaped notion of consistency, where each occurrence is tied directly to the same reference rather than only to local visual context, as illustrated in Figure~\ref{fig:motivation}. Grounding every entity back to its canonical reference, rather than to the most recent generated appearance, is therefore essential for preventing cumulative divergence.

This motivates a solution centered on \textbf{entity-level visual memory}: building and maintaining references for each entity—characters, objects, and locations—so that later shots can use reliable references for the entities they contain. Such memory must satisfy three properties.
First, it should be \emph{high-quality}: admitted references should depict entities clearly and unambiguously so that their identity cues can reliably guide later generation.
Second, it should enforce \emph{canonical reference prioritization}: once a clear canonical reference is available, it should be protected in memory and used by default for later shots, so that all occurrences remain tied to the same identity standard.
Third, it should support \emph{dynamic maintenance}: while keeping the canonical reference fixed, the memory should admit a small set of canonical-consistent supplementary references that cover what the canonical reference does not, such as character expression or view changes and location extensions across camera directions or newly visible areas.

Based on these observations, we present \textbf{GroundShot}, a training-free, model-agnostic agentic framework for multi-shot video generation. GroundShot builds entity-level visual memory from accepted generated shots through grounding, and uses quality-aware shot scheduling to generate stronger reference-source shots before dependent shots.
Since entity references are self-generated, not every grounded entity is suitable as a reference—a partially occluded face, a motion-blurred figure, or a character seen only from behind carries insufficient or misleading identity information. GroundShot therefore filters out unreliable grounded references before admitting them into memory.

To this end, GroundShot combines three key mechanisms.
\emph{Quality-aware shot scheduling} identifies which entities appear in which shots and which shots are most likely to provide clear references, then constructs a generation order that prioritizes those shots while deferring shots predicted to yield low-quality references.
After each shot is generated, \emph{entity-level grounding} extracts candidate entity references from the output, and quality checks retain only candidates reliable enough to guide later generation; accepted references are admitted into entity-level visual memory.
When generating a later shot, \emph{reference retrieval} uses the canonical reference by default and selects a canonical-consistent supplementary reference only when it better covers the target shot's expression, view, or location requirement. GroundShot then conditions generation on these selected entity-level references.
Together, these mechanisms ensure that reliable references are acquired, maintained, and reused systematically rather than inherited from narrative order.

\vspace{0.5em}
\noindent\textbf{Contributions.} We make the following contributions:

\begin{enumerate}
    \item We introduce \textbf{GroundShot}, a training-free and model-agnostic agentic framework for visually consistent multi-shot video generation. Unlike sequential frame-level conditioning, GroundShot formulates multi-shot consistency as an agentic reference-management problem, explicitly managing entity-level visual memory and decoupling shot generation order from narrative order through quality-aware shot scheduling.

    \item We introduce \textbf{GroundBench}, a diagnostic benchmark that evaluates multi-shot consistency at the entity level and organizes scripts into controlled challenge modules for fine-grained failure analysis.

    \item Extensive experiments on GroundBench show that GroundShot outperforms strong multi-shot baselines while requiring no additional training or model modification, and ablations confirm the critical role of each agentic component.
\end{enumerate}

% ============================================================================
\section{Related Work}

\paragraph{Multi-Shot and Long-Form Video Generation.}
Text-to-video models have advanced from GAN-based~\cite{text2video_zero} and diffusion-based~\cite{blattmann2023align,animatediff} systems to DiT architectures~\cite{sora,yang2024cogvideox,wan2024wan,moviegen} capable of photorealistic single-shot clips, but extending them to multi-shot narratives with cross-shot consistency remains open.
Architecture-level methods embed consistency into the backbone through shared attention~\cite{storydiffusion}, streaming architectures~\cite{shotstream}, dual-masking~\cite{mask2dit}, character-aware transitions~\cite{infinitystory}, explicit memory modules~\cite{storymem,videomemory,onestory}, cache-guided autoregressive diffusion~\cite{filmweaver}, or holistic multi-shot generation~\cite{holocine}.
Pipeline-level methods keep the backbone frozen and address consistency through LLM-planned scripts~\cite{videostudio}, entity-level layout planning~\cite{multishotmaster}, storyboard anchoring~\cite{stage}, controllable shot transitions~\cite{shotdirector,shotadapter}, chain-of-thought reasoning~\cite{videogenofthought}, hierarchical autoregressive-diffusion generation~\cite{moviedreamer}, cross-shot face consistency~\cite{echoshot}, content anchors~\cite{gloria}, arbitrary identity references~\cite{anyid}, or end-to-end cinematic production~\cite{vlogger,captaincinema}.
Among these approaches, memory-based methods~\cite{storymem,videomemory,a2rd} are most closely related to ours, maintaining a visual memory that persists across shots to mitigate drift. However, their memories store and retrieve entire frames, delegating entity disentanglement to the generation model, and their generation order remains tied to narrative sequence. GroundShot instead operates at the entity level---grounding, verifying, and selecting per-entity references while decoupling generation order from narrative order.

\paragraph{Reference-Conditioned and Subject-Driven Generation.}
Identity-preserving generation conditions on reference images via cross-attention adapters~\cite{ipadapter,photomaker} for images, and reference-conditioned video generation models~\cite{phantom,dreamvideo,libragen} for video.
Closed-source systems including Gen-3~\cite{gen3}, Kling~\cite{kling}, Seedance~\cite{seedance}, Vidu~\cite{vidu}, Luma Dream Machine~\cite{lumadreammachine}, and Hailuo~\cite{hailuo} also support reference-conditioned video with high fidelity.
These methods already achieve strong within-shot visual consistency given references, but do not address cross-shot coordination: which references to use, where they come from, and when to acquire them.
GroundShot builds on this capability---it wraps any compatible T2V or reference-conditioned backbone with an agentic layer that manages references across the full multi-shot generation.

\paragraph{Agentic and LLM-Driven Video Systems.}
LLMs have been used as planners for video generation through structured scene layouts~\cite{videodirectorgpt,llmgroundeddiffusion,llmgroundedvideo}.
Multi-agent systems further decompose video production into specialized roles for scripting, storyboarding, and animation~\cite{storyagent,dreamfactory,cameraartist,automatedmovie,filmagent,anime,mmstoryagent}.
Test-time self-improvement~\cite{vista} and closed-loop cognitive orchestration~\cite{muse} introduce feedback into generation, while concurrent work~\cite{agenticvideogen} proposes executable event graphs with tool-constrained planning.
However, these systems primarily use agents for \emph{planning}---constructing prompts, layouts, or workflows---rather than for actively managing the generation process itself.
GroundShot employs agentic reasoning to actively manage the generation process---scheduling shot synthesis order, grounding and verifying entities from generated outputs, and maintaining a curated visual memory across shots.

\paragraph{Visual Grounding.}
Open-vocabulary grounding has matured through grounded pre-training~\cite{groundingdino}, spatial grounding for generation~\cite{gligen}, and unified referring expression models~\cite{liang2025referdino}.
These are typically used as perception modules or spatial control signals.
GroundShot repurposes grounding as a feedback mechanism: entity crops extracted from generated videos become conditioning inputs for subsequent shots, and predicted grounding quality drives the scheduling policy.

\paragraph{Concurrent Work.}
Several concurrent efforts also address multi-shot or long-form consistency.
On the evaluation side, EntityBench~\cite{entitybench} independently proposes entity-level benchmarking and pairs it with a per-entity memory as a generation baseline.
On the generation side, A$^2$RD~\cite{a2rd} employs agentic refinement for multi-shot synthesis; CausalCine~\cite{causalcine} extends autoregressive generation with shot-boundary awareness; IAMFlow~\cite{iamflow} tracks entity identities across prompt transitions via LLM-extracted textual attributes; and Echo-Forcing~\cite{echoforcing} manages scene-level KV-cache memory for interactive long video.
GroundShot shares the goal of improving long-form consistency, but instead of propagating identity through a drifting sequence, it takes an agentic memory-management approach: using quality-aware scheduling to establish high-quality canonical references for recurring entities and grounding later appearances against the same reference standard.

\section{Method}

\subsection{Overview}

Given a text script $\mathcal{S}=\{s_1,\ldots,s_N\}$ and an optional global caption $\mathcal{C}$, GroundShot generates a video sequence $\mathcal{V}=\{v_1,\ldots,v_N\}$.
To support cross-shot consistency, GroundShot uses an \emph{entity-level visual memory} $\mathcal{R}$.
For each recurring entity---character, object, or location---this memory stores high-quality references, keeping that entity's canonical reference protected, and adds reliable supplementary references as generation proceeds.

GroundShot first analyzes recurring entities in the script and predicts which shots can provide reliable references for them. The scheduler then uses these predictions to determine a generation order that may differ from the narrative order, prioritizing shots that establish stronger references for later use; see Section~\ref{sec:scheduling}.

\begin{figure*}[t]
    \centering
    \includegraphics[width=\textwidth]{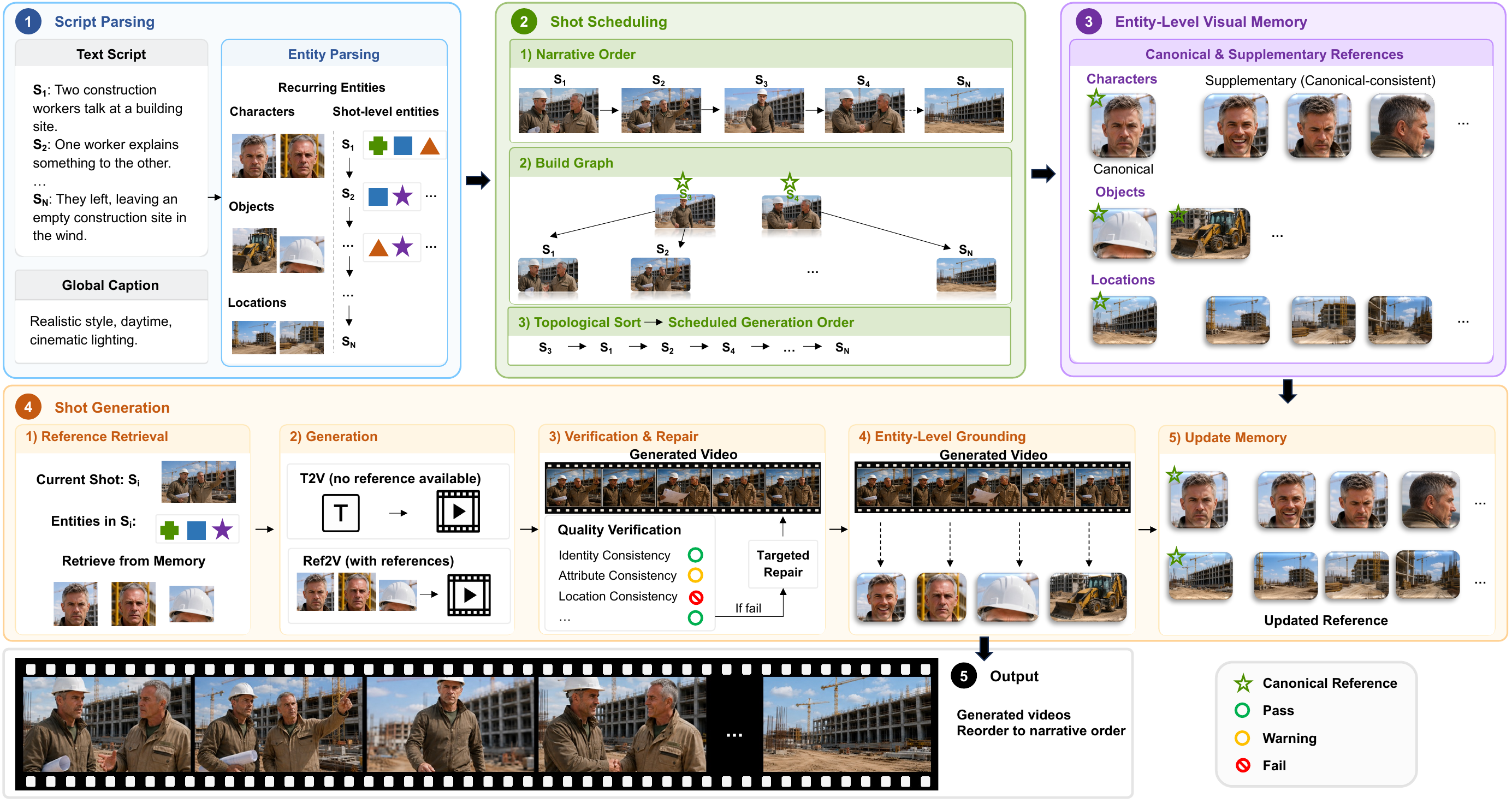}
    \caption{\textbf{Overview of GroundShot.}
    \textbf{(1) Script parsing} extracts recurring characters, objects, and locations, and identifies the entity set for each shot.
    \textbf{(2) Shot scheduling} builds a dependency graph from predicted reference-source shots and topologically sorts it into a scheduled generation order.
    \textbf{(3) Entity-level visual memory} stores canonical and canonical-consistent supplementary references for characters, objects, and locations.
    \textbf{(4) Shot generation} retrieves references for the current shot, chooses T2V or Ref2V generation, verifies and repairs the result, grounds entity-level evidence from the generated video, and updates visual memory with reliable references.}
    \label{fig:overview}
\end{figure*}

At each scheduled shot, GroundShot retrieves usable references from visual memory: shots without usable references are generated by T2V, while available entity references enable Ref2V. The result is verified, repaired if needed, and grounded into entity crops or scene references; reliable grounded outputs are admitted into visual memory, and the recorded generation experience guides later decisions. Once all shots have been generated, the clips are sorted back to narrative order; Figure~\ref{fig:overview} illustrates the main pipeline stages, and Algorithm~\ref{alg:groundshot} summarizes the full flow.

\begin{algorithm}[t]
\caption{GroundShot}
\label{alg:groundshot}
\begin{algorithmic}[1]
\REQUIRE Script $\mathcal{S}=\{s_1,\ldots,s_N\}$, optional global caption $\mathcal{C}$
\ENSURE Videos $\mathcal{V}=\{v_1,\ldots,v_N\}$ in narrative order
\item[\textbf{Notation:}] $\mathcal{E}_i$: entity set; $p_i$: prompt; $\pi$: scheduled order; $\eta_i$: experience advice; $\mathcal{I}_i$: references; $m_i$: generation mode; $\rho_i$: feedback; $\mathcal{D}_i$: candidate reference
\STATE $\{\mathcal{E}_i,p_i\}_{i=1}^{N} \leftarrow \operatorname{ParseScript}(\mathcal{S},\mathcal{C})$
\STATE $\pi\leftarrow \operatorname{QualityAwareSchedule}(\{s_i,\mathcal{E}_i\}_{i=1}^{N})$
\STATE Initialize entity-level visual memory $\mathcal{R}_0\leftarrow\emptyset$
\STATE Load generation experience $\mathcal{X}$
\FOR{$t=1$ to $N$}
    \STATE $i\leftarrow \pi_t$
    \STATE $\eta_i\leftarrow\operatorname{RetrieveExperience}(\mathcal{X},s_i,\mathcal{E}_i,p_i)$
    \STATE $\mathcal{I}_i\leftarrow\operatorname{RetrieveReferences}(\mathcal{R}_{t-1},\mathcal{E}_i,s_i)$
    \STATE $m_i\leftarrow\operatorname{SelectGenerator}(\mathcal{I}_i,\mathcal{E}_i,\eta_i)$
    \STATE $\tilde{v}_i\leftarrow\mathcal{G}_{m_i}(p_i,\mathcal{I}_i)$
    \STATE $v_i,\rho_i\leftarrow\operatorname{VerifyAndRepair}(\tilde{v}_i,p_i,\mathcal{I}_i,\mathcal{E}_i,\eta_i)$
    \STATE $\mathcal{D}_i\leftarrow\operatorname{EntityLevelGrounding}(v_i,\mathcal{E}_i)$
    \STATE $\mathcal{R}_{t}\leftarrow\operatorname{UpdateVisualMemory}(\mathcal{R}_{t-1},\mathcal{D}_i)$
    \STATE $\mathcal{X}\leftarrow\operatorname{RecordExperience}(\mathcal{X},s_i,p_i,m_i,\rho_i,\mathcal{D}_i)$
\ENDFOR
\RETURN $\{v_i\}_{i=1}^{N}$ sorted by narrative shot ID
\end{algorithmic}
\end{algorithm}

\subsection{Quality-aware shot scheduling}
\label{sec:scheduling}

GroundShot uses quality-aware shot scheduling to determine the generation order. The goal is to generate reference-source shots for recurring entities before generating the remaining shots in which those entities appear.

\paragraph{Script parsing.}
An LLM parser processes the full script and optional global caption $\mathcal{C}$ jointly. It builds a global set of entities, types each entity as a character, object, or location, and links different textual mentions across shots so that the same character, object, or location receives the same entity ID. For each shot $s_i$, the parser outputs the set $\mathcal{E}_i$ of entities appearing in that shot. It also assigns an entity priority for later scheduling: characters receive the highest priority, objects come next, and locations are prioritized after them. After parsing, the scheduler knows where each recurring entity appears. It can then compare those appearances and decide which shot should be generated first to obtain a reliable reference for that entity.

\paragraph{Shot scheduling.}
A recurring entity may appear in multiple shots, but these appearances are not equally useful as future references. For example, a character occupying only a few pixels in a wide establishing shot gives a much weaker identity signal than the same character shown in a medium close-up. If generation follows narrative order, an early weak appearance may become the reference. Because this reference contains little usable identity evidence, later shots conditioned on it receive a weak constraint and struggle to preserve identity consistently.
GroundShot therefore treats reference reliability as a scheduling objective: a shot should be generated early when it is expected to provide a reliable visual reference for entities that reappear elsewhere, not simply because it appears early in the story. For each recurring entity $e$ and each shot $s_i$ where it appears, GroundShot predicts a reference-quality score:
\begin{equation}
    q_{\mathrm{src}}(s_i,e)=Q_{\mathrm{ref}}(s_i,e;\mathcal{C}),
\end{equation}
where $q_{\mathrm{src}}(s_i,e)$ measures how useful shot $s_i$ is expected to be as an early reference source for entity $e$. $Q_{\mathrm{ref}}$ is a reference-quality estimator that uses the shot text $s_i$, the parsed entity description for $e$, and the global caption $\mathcal{C}$ to assess how clearly and completely $e$ is expected to appear. For each recurring entity, the shot with the highest $q_{\mathrm{src}}(s_i,e)$ is selected as the source shot.
The selected source shot is then constrained to be generated before the other shots in which the same entity appears.

GroundShot represents all such constraints as a directed graph over shots. For each recurring entity, directed edges point from its selected source shot to the other shots where that entity appears. Constraints from different entities are not always mutually consistent: different entities can prefer opposite orders over overlapping shots, which creates cycles. The scheduler resolves cycles before sorting the graph. When a cycle is detected, it preserves the constraints that better serve reference construction, prioritizing higher-priority entities, larger predicted reference-quality gains, and then orders closer to narrative order. After all cycles are removed, a topological sort of the resulting directed acyclic graph (DAG) gives the generation order $\pi$. Thus, GroundShot decouples the order in which shots are generated from the order in which they are finally viewed.

\subsection{Entity-Level Visual Memory}
\label{sec:visual_memory}

Given the scheduled generation order, GroundShot needs a way to turn early reference-source shots into reusable conditioning signals for later shots.
Unlike frame-level memory that stores and retrieves entire frames---requiring the generation model to implicitly disentangle characters, objects, and background---the entity-level visual memory $\mathcal{R}$ organizes references separately for each entity. It stores foreground crops for characters and objects, and scene references for locations. For an entity $e$, the memory contains a protected canonical reference $r^*_e$ and, when available, supplementary references that capture useful appearance variation. The memory can be initialized with user-provided references; when only text is provided, it is built from GroundShot's accepted generated videos.

\paragraph{Entity-level grounding.}
After a generated shot $v_i$ is accepted by the shot generation process in Section~\ref{sec:adaptive_gen},
GroundShot performs entity-level grounding: it samples frames and uses each entity's parsed name and description to identify matching regions for characters and objects.
Each localized region becomes a candidate crop for entity $e$.

Location grounding requires a separate path because, unlike characters and objects, a location cannot be represented by directly cropping a foreground region from the generated frame. GroundShot grounds all foreground entities, forms their union mask, and sends the frame with the mask to an instruction-following image editing model, which removes the masked foreground regions and reconstructs the scene behind them.
The edited image is treated as a candidate scene reference for the location entity.

Before adding any grounded candidate into memory, GroundShot applies type-specific quality checks, including image-quality metrics and a VLM-based semantic check.
This is necessary because weak references can propagate errors: a character that is partially occluded, motion-blurred, or seen only from behind may carry insufficient identity information; an object that is small, blurred, or occluded may provide unreliable shape or texture cues; and a flawed location reconstruction can introduce artifacts or leave visible human figures from the removed foreground that later shots may reproduce.
GroundShot therefore computes a type-aware candidate quality score $q_{\mathrm{cand}}(c,e)$ for each candidate reference $c$: character candidates are evaluated for face visibility, frontality, and sharpness; object candidates for recognizable and unoccluded foreground evidence; and location candidates for whether the reconstructed scene is safe to reuse.
A candidate is admitted only if
\begin{equation}
    q_{\mathrm{cand}}(c,e)\geq\tau_{\mathrm{type}(e)},
\end{equation}
where $\tau_{\mathrm{type}(e)}$ is a type-specific threshold. Candidates below this threshold are discarded and are not written into visual memory.

\paragraph{Visual memory dynamic maintenance.}
For each entity, the first sufficiently strong reference becomes the canonical reference. Later candidates do not overwrite it. They are considered only as supplementary references after passing the type-aware quality filter above. A new candidate can be added only if it remains consistent with the canonical reference and contributes non-redundant visual evidence:
\begin{equation}
    \operatorname{sim}(c,r^*_e)\geq\theta_{\mathrm{id}}
    \quad\wedge\quad
    \max_{r\in\mathcal{R}_e}\operatorname{sim}(c,r)\leq\theta_{\mathrm{div}},
\end{equation}
where $\mathcal{R}_e$ is the current reference set for entity $e$, $\theta_{\mathrm{id}}$ is the minimum canonical-consistency threshold, and $\theta_{\mathrm{div}}$ is the maximum allowed similarity to existing references for diversity. The identity gate prevents drift away from the canonical reference, while the diversity gate keeps only references that add useful variation rather than near-duplicates. For characters, such supplementary references mainly help with facial expressions or large view changes; and for locations, they cover the same environment from different camera directions or newly expanded regions that spatially extend the location along the camera or character trajectory, rather than forcing every location-conditioned shot to reuse the first local background view.
Low-quality, inconsistent, or redundant candidates are discarded, so visual memory dynamic maintenance keeps only a compact set of canonical-consistent supplementary references.

\subsection{Shot Generation}
\label{sec:adaptive_gen}

Given the generation order $\pi$ computed by the scheduler, GroundShot generates shot $s_i$ with $i=\pi_t$ at step $t$.
GroundShot first builds a layered prompt $p_i$ by combining the overall style, setting, lighting, and color tone from the global caption, the parsed descriptions of the entities in $s_i$, and the shot prompt. Rather than appending the full global caption to every shot, GroundShot uses it only for shared style and scene-level cues, while the entity layer contains only the entities parsed for $s_i$; this prevents entities that appear elsewhere in the script from being introduced into the current shot through the global prompt.
GroundShot then retrieves available references from visual memory, selects either T2V or Ref2V generation, verifies the result, and uses accepted shots to extract new entity references for visual memory update as described in Section~\ref{sec:visual_memory}. In parallel, GroundShot maintains generation experience $\mathcal{X}$ that records what generation and repair choices worked in earlier shots.

\paragraph{Reference retrieval.}
For the current shot, let $\mathcal{I}_i$ denote the references retrieved from $\mathcal{R}_{t-1}$ for entities in $\mathcal{E}_i$.
References are retrieved independently for each entity. GroundShot uses the canonical reference by default so that different shots follow the same identity standard.
When canonical-consistent supplementary references are available, GroundShot can append one of them to better match the target shot while still keeping the canonical reference as the identity anchor.
To decide whether a supplementary reference is useful, the retriever, a VLM, scores auxiliary references for target-shot compatibility, considering the shot prompt, expected framing, viewing direction, action or expression, and location coverage, and then chooses whether to add the best auxiliary reference.

Reference availability determines the generation mode. If no usable foreground reference is available, GroundShot bootstraps the shot with a text-to-video generator (T2V). This path lets GroundShot construct visual memory from text scripts without requiring external reference images. If at least one foreground entity has a usable reference, GroundShot uses a reference-conditioned video generator (Ref2V) with the available references. Because memory is built online, $\mathcal{I}_i$ may cover only a subset of $\mathcal{E}_i$; entities without references remain specified by the layered prompt. For locations without scene references, the global and shot-level prompt layers carry the location information until a later shot provides reliable visual evidence. Let $m_i$ denote the selected generation mode and $\mathcal{G}_{m_i}$ the corresponding generator:
\begin{equation}
    \tilde{v}_i=\mathcal{G}_{m_i}(p_i,\mathcal{I}_i),\quad
    m_i\in\{\mathrm{T2V},\mathrm{Ref2V}\}.
\end{equation}

\paragraph{Shot verification.}
After generating a candidate video $\tilde{v}_i$, GroundShot verifies it before keeping it as $v_i$ and allowing it to contribute new references to visual memory. A VLM critic samples frames from $\tilde{v}_i$ and compares them with the shot description, selected references, and expected entity set. Instead of returning only a scalar similarity, the critic reports structured issues such as entity count errors, identity mismatch, clothing mismatch, style drift, lighting inconsistency, unnatural pose, and general quality degradation. A shot passes when its quality score is above threshold and no severe issue is present:
\begin{equation}
    \operatorname{pass}(\tilde{v}_i)=
    \bigl(Q(\tilde{v}_i)\geq\tau_{\mathrm{pass}}\bigr)
    \wedge \neg\operatorname{Severe}(\tilde{v}_i).
\end{equation}
Here $Q(\tilde{v}_i)$ is the scalar quality score assigned by the VLM critic, $\tau_{\mathrm{pass}}$ is the minimum acceptance threshold, and $\operatorname{Severe}(\tilde{v}_i)$ is a Boolean predicate indicating whether the critic reports any severe issue, such as identity mismatch, missing entities, or major motion/rendering artifacts.
If the shot fails, GroundShot tries a small number of targeted fixes according to the reported issue type. For example, identity mismatch can trigger stronger reference conditioning or a different retrieved reference; missing or extra entities can add explicit count and presence instructions to the revised prompt; style or lighting drift can add a stronger reminder of the global prompt's related descriptions; severe motion or rendering artifacts can trigger a new seed, etc. If no attempt passes within the retry budget, GroundShot keeps the highest-scoring attempt as $v_i$.

\paragraph{Generation experience.}
Generation experience $\mathcal{X}$ records what happened in previous generation attempts, especially when a shot fails verification or requires repair. Each record stores which references were used, the generation parameters, the failure issues reported by the critic, the repair actions attempted, whether each repair succeeded, the final verification score, etc.
Before generating a new shot, GroundShot queries $\mathcal{X}$ using the current shot's parsed entities and shot prompt, then uses records with similar settings to form guidance $\eta_i$.
After generation, if verification reports a failure, GroundShot orders repair actions using previous successes and failures for the same issue type.
After the shot is completed, its final result is recorded back into $\mathcal{X}$, allowing later shots and future runs to benefit from past records.

% ============================================================================
\section{Experiments}

\subsection{GroundBench: Multi-Shot Consistency Benchmark}

Recent multi-shot benchmarks have begun to evaluate long-range coherence and story-level consistency, but most remain holistic: they do not localize failures to individual recurring entities, nor isolate the challenge dimension that triggered the failure. As a result, they can show that a method is inconsistent, but not whether the error comes from a particular character, object, or location, or from factors such as viewpoint change, occlusion, multi-identity confusion, or scene revisits. We construct \textbf{GroundBench}, a diagnostic benchmark with both properties. GroundBench is \emph{entity-grounded}: metrics are computed on entity level rather than only full frames, allowing failures to be attributed to specific recurring entities---characters, objects, or locations. It is also \emph{modular}: scripts are organized into diagnostic modules that systematically vary one challenge dimension at a time, enabling fine-grained failure analysis.

\paragraph{Benchmark design.}
GroundBench contains \textbf{54 multi-shot scripts} ($\sim$300 shots total) spanning four diagnostic axes: \emph{identity preservation} under scale/lighting/viewpoint/occlusion changes, \emph{multi-identity discrimination} with visually distinct or similar characters, \emph{structural patterns} such as scene revisits, intercut timelines, long sequences, and late character introduction, and \emph{robustness probes} involving style shifts, crowds, and pronoun-heavy coreference. Each axis is further divided into controlled sub-categories with three difficulty levels; the detailed taxonomy is provided in the supplementary material.

\paragraph{Metrics.}
GroundBench reports diagnostic metrics designed for multi-shot consistency.
Our primary entity-level metrics are: \textbf{ARC} (Anchor-Relative Consistency), which measures whether each recurring entity remains consistent with its first valid appearance; \textbf{WCI} (Worst-Case Identity), which measures the lowest similarity between an entity's first valid appearance and any later occurrence; \textbf{XCD} (Cross-Character Discrimination), which measures the separation between same-character and different-character similarities in multi-identity scenes; and \textbf{EAF} (Entity-Attribute Fidelity), which measures whether each detected entity matches its textual appearance description. We also report \textbf{ESC} (Environment/Scene Consistency) to evaluate whether recurring settings remain visually coherent across shots after foreground entities are masked out.
As a shot-level consistency baseline, we report \textbf{CSC} (Cross-Shot Consistency): the mean pairwise cosine similarity of ViCLIP~\cite{wang2023internvid_viclip} video features across all shot pairs, following prior works~\cite{holocine,multishotmaster,storymem}.
All metrics are computed with a method-independent Unified Evaluation Detector (UED).
For video quality assessment, we additionally report aesthetic quality and dynamic degree from VBench~\cite{huang2024vbench_comprehensive}. Formal definitions are provided in the supplementary material.
% ============================================================================
\subsection{Experimental Setup}

We compare against four multi-shot baselines:
\textbf{HoloCine}~\cite{holocine}, \textbf{MultiShotMaster-1.3B}/\textbf{MultiShotMaster-14B}~\cite{multishotmaster} and \textbf{StoryMem}~\cite{storymem}.
For the experiments reported here, GroundShot uses Vidu~\cite{vidu} as the primary video generation backend for both bootstrap T2V shots and reference-conditioned shots (\texttt{viduq3-pro} and \texttt{viduq3-mix}).
To evaluate backend-agnosticism, we additionally report results with Kling~\cite{kling} as an alternative commercial backend and Wan2.1-T2V~\cite{wan2024wan} + Phantom-14B~\cite{phantom} as a fully open-source backend, all under the same agentic framework.
We use 4-second, 16:9, 720p clips with automatic motion amplitude and audio disabled.
Entity parsing uses GPT-4o~\cite{gpt4o}. GPT-4.1~\cite{gpt41} is used for the LLM-based decisions in scheduling, reference retrieval, and shot verification, such as reference-quality estimation, target-shot reference scoring, and VLM critique.
Entity-level grounding uses ReferDINO~\cite{liang2025referdino} to localize foreground entity candidates. Visual-memory admission follows the type-aware quality checks in Section~\ref{sec:visual_memory}, with a minimum candidate quality of 0.4 and a character face-confidence threshold of 0.3; a candidate becomes the canonical reference when its quality score is at least 0.85.

% ============================================================================
\subsection{Comparison}

\begin{table*}[t]
\centering
\footnotesize
\caption{Quantitative comparison on GroundBench. Best in \textbf{bold}, second-best \underline{underlined}. $\uparrow$: higher is better.}
\label{tab:main_results}
\vspace{0.3em}
\setlength{\tabcolsep}{4pt}
\begin{tabular*}{\textwidth}{@{\extracolsep{\fill}}l|ccccc|cc|cc}
\toprule
\multirow{2}{*}{\textbf{Method}} & \multicolumn{5}{c|}{\textbf{Entity-Level}} & \multicolumn{2}{c|}{\textbf{Shot-Level}} & \multicolumn{2}{c}{\textbf{Quality}} \\
\cmidrule(lr){2-6}\cmidrule(lr){7-8}\cmidrule(lr){9-10}
& \textbf{ARC}$\uparrow$ & \textbf{XCD}$\uparrow$ & \textbf{WCI}$\uparrow$ & \textbf{EAF}$\uparrow$ & \textbf{ESC} & \textbf{FaceSim}$\uparrow$ & \textbf{CSC}$\uparrow$ & \textbf{Aesth.}$\uparrow$ & \textbf{Dyn.} \\
\midrule
HoloCine~\cite{holocine}                    & 0.368 & 0.119 & 0.212 & 0.617 & 0.549 & 0.417 & 0.621 & 0.542 & 0.539 \\
MultiShotMaster-1.3B~\cite{multishotmaster} & 0.220 & 0.057 & 0.163 & 0.613 & 0.480 & 0.271 & 0.509 & 0.521 & 0.625 \\
MultiShotMaster-14B~\cite{multishotmaster}  & 0.282 & 0.078 & 0.204 & 0.611 & 0.397 & 0.367 & 0.572 & 0.540 & 0.375 \\
StoryMem~\cite{storymem}                    & 0.314 & 0.125 & 0.240 & 0.616 & \textbf{0.700} & 0.368 & 0.656 & 0.536 & \textbf{0.750} \\
\midrule
Vidu-Naive                                  & 0.346 & 0.114 & 0.221 & 0.615 & 0.565 & 0.404 & 0.635 & 0.545 & 0.615 \\
\textbf{GroundShot (Vidu)}                         & \textbf{0.426} & \textbf{0.157} & \textbf{0.264} & \textbf{0.634} & 0.670 & \textbf{0.449} & \textbf{0.671} & \underline{0.552} & 0.640 \\
\midrule
Kling-Naive                                 & 0.355 & 0.116 & 0.227 & 0.612 & 0.583 & 0.408 & 0.642 & 0.550 & 0.620 \\
\textbf{GroundShot (Kling)}                        & \underline{0.414} & \underline{0.151} & \underline{0.255} & \underline{0.629} & \underline{0.688} & \underline{0.441} & \underline{0.663} & \textbf{0.558} & \underline{0.715} \\
\midrule
Phantom-Naive                               & 0.301 & 0.081 & 0.198 & 0.598 & 0.526 & 0.356 & 0.586 & 0.528 & 0.580 \\
\textbf{GroundShot (Phantom)}                      & 0.376 & 0.127 & 0.235 & 0.614 & 0.634 & 0.402 & 0.642 & 0.534 & 0.610 \\
\bottomrule
\end{tabular*}
\end{table*}

\paragraph{Quantitative results.}
Table~\ref{tab:main_results} reports the main quantitative comparison. GroundShot achieves the best entity-level consistency on ARC, XCD, WCI, and EAF. The gains are most pronounced on ARC and XCD, indicating that quality-aware shot scheduling establishes stronger first-appearance anchors, while entity-level grounding better separates multiple recurring identities.
GroundShot improves ESC over its same-backend Naive variants, showing that the proposed memory and scheduling framework also benefits scene revisits. However, StoryMem obtains the highest ESC, slightly above the Kling-backed GroundShot variant. This is not contradictory to its weaker ARC/WCI: ESC compares background DINOv2 embeddings after masking foreground characters and objects, so a whole-frame/keyframe memory mechanism can preserve room layout, lighting, materials, and overall visual atmosphere even when character faces, clothing, or identities drift.
The contrast among identity-level and shot-level metrics is important. StoryMem obtains strong CSC and ESC scores, suggesting good holistic and background-level cross-shot coherence, but its ARC and WCI remain substantially lower than GroundShot. HoloCine exhibits a similar pattern: its holistic attention improves overall coherence, but does not guarantee per-entity fidelity. GroundShot improves both entity-level and shot-level consistency, suggesting that grounding references at the entity level strengthens fine-grained identity without disrupting global visual coherence.
Quality metrics remain comparable across methods. GroundShot stays within the range of strong baselines on aesthetic quality and dynamic degree, improving consistency without sacrificing visual quality or dynamics.

\paragraph{Diagnostic failure modes.}
GroundBench's diagnostic axes make the weaknesses of each baseline more explicit. On \emph{identity preservation}, the same entity reappears under scale, lighting, angle, temporal-gap, and occlusion changes. HoloCine~\cite{holocine}'s holistic attention can keep shots stylistically coherent, but fine-grained identity cues may be diluted when the entity changes scale or view. StoryMem~\cite{storymem}'s keyframe memory can be similarly brittle when the remembered frame is wide, occluded, or pose-mismatched, preserving surrounding context while facial or clothing details drift; ARC and WCI expose this gap. On \emph{multi-identity discrimination}, including visually similar pairs, character groups, character--object binding, and appearance-swap stress, MultiShotMaster~\cite{multishotmaster} is especially limited: controllable shot/reference injection helps specify the scene, but does not by itself resolve which recurring visual evidence belongs to which character or object, yielding low XCD and attribute leakage. On \emph{structural pattern}, such as gallery--office--gallery revisits, narrative-order and frame-level memories are vulnerable to early low-quality anchors, memory aging, and background-preserving but entity-drifting revisits; this explains why StoryMem can score well on CSC while remaining weaker on ARC/WCI. On \emph{robustness probes}, baselines often keep the clip plausible but lose the target entity, rewriting its identity under the new style, locking onto a nearby person, or drifting when the prompt no longer names the character.
Thus, GroundBench turns aggregate consistency into a diagnosis of which ability is missing: stable single-entity identity, cross-identity separation, long-range structure, or entity binding under ambiguity.

\paragraph{Same-backend comparison.}
Since GroundShot uses external video generation backends while baselines use their own built-in generation components, a natural concern is that GroundShot's gains may come from a stronger video generation backend rather than the proposed agentic framework. To address this concern, we evaluate GroundShot with three backends---Vidu~\cite{vidu} (commercial), Kling~\cite{kling} (commercial), and Wan2.1~\cite{wan2024wan}+Phantom~\cite{phantom} (fully open-source)---and pair each with a \textbf{Naive} variant that uses the same backend but generates shots in narrative order, simply using the last frame containing a visible face from the preceding shot as the reference image.
As Table~\ref{tab:main_results} shows, adding GroundShot consistently improves entity-level metrics over the corresponding Naive variant across all three backends. This pattern supports that the improvement comes from our framework rather than from a particular generation backend. The open-source Wan2.1+Phantom setting further indicates that the framework is not tied to commercial backends: even with a weaker open-source backend, the same grounded entity-level memory and shot scheduling yield gains.

\begin{figure*}[t]
    \centering
    \includegraphics[width=0.95\textwidth]{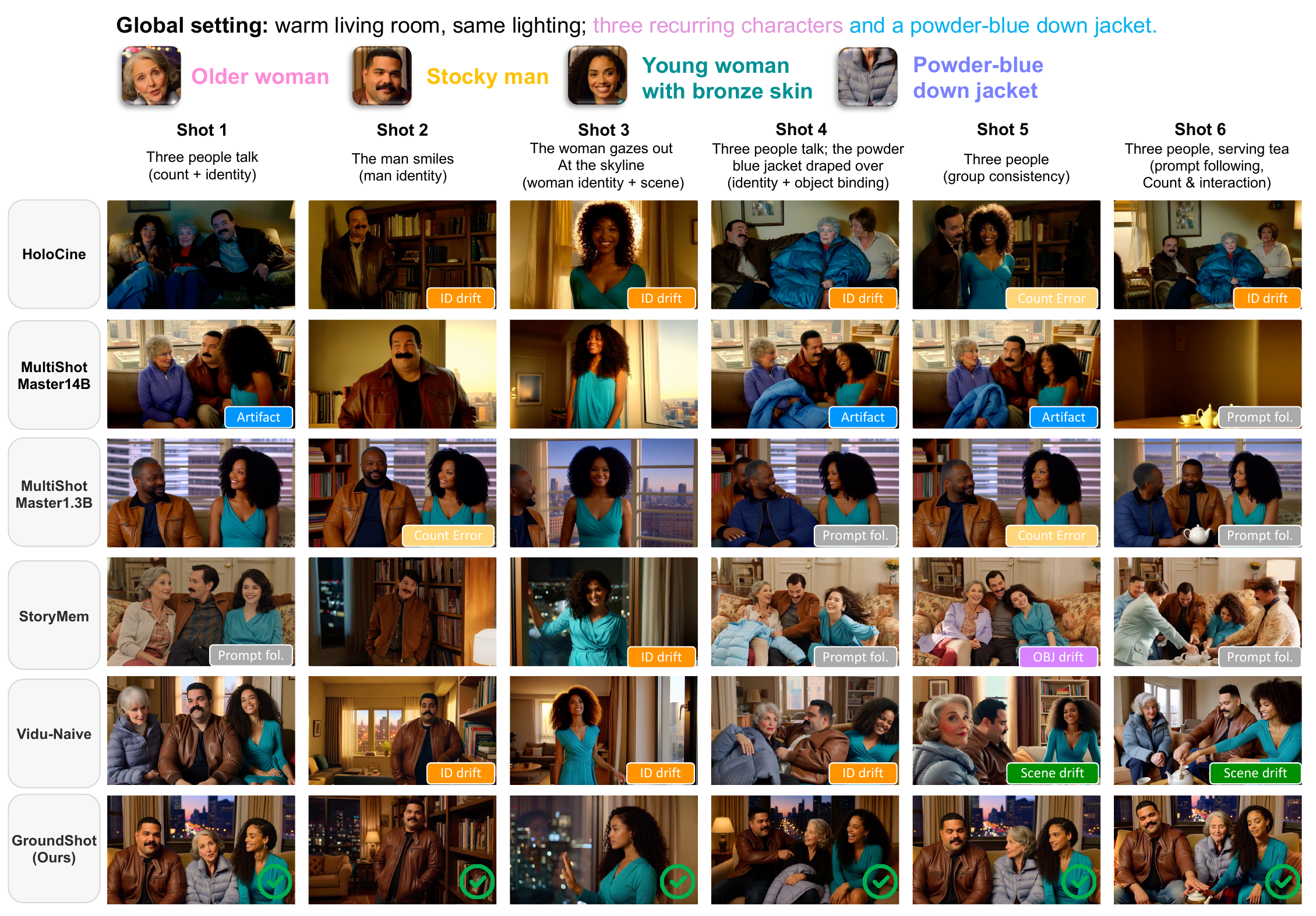}
    \caption{Qualitative comparison on a six-shot living-room script. The setting contains three recurring characters and a powder-blue down jacket, testing identity/object/scene consistency, character-count correctness, and prompt following. GroundShot preserves the older woman, stocky man, young woman, and jacket across individual and group shots while following the requested actions. Baselines exhibit identity drift, count errors, artifacts, prompt-following failures, object drift, or scene drift.}
    \label{fig:qualitative_comparison}
\end{figure*}

\paragraph{Qualitative comparison.}
Figure~\ref{fig:qualitative_comparison} presents a six-shot living-room script with three recurring characters and a powder-blue down jacket. The example stresses several entity-level requirements at once: preserving each person's identity across individual and group shots, maintaining the correct character count, binding the jacket to the older woman, and following shot-specific actions. Baselines fail in different ways. HoloCine often drifts identity or count despite a coherent overall style; MultiShotMaster variants introduce artifacts, count errors, or prompt-following failures; StoryMem can preserve a plausible room layout while drifting identities or object ownership; and Vidu-Naive shows both identity and scene drift. GroundShot maintains the three characters, the jacket binding, the warm living-room setting, and the requested interactions across all six shots.

% ============================================================================
\subsection{Ablation Studies}

We ablate the key components of GroundShot to isolate their contributions. All ablations are evaluated on the full GroundBench with identical scripts and seeds.

\begin{table}[t]
\centering
\caption{Ablation study on GroundBench. Each row removes one component from the full system. Entity grounding and scheduling are the two most critical components, together accounting for the majority of GroundShot's gains over baselines.}
\label{tab:ablation}
\vspace{0.3em}
\footnotesize
\setlength{\tabcolsep}{3pt}
\begin{tabular*}{\columnwidth}{@{\extracolsep{\fill}}L{0.36\columnwidth}|ccc|c}
\toprule
\textbf{Configuration} & \textbf{ARC}$\uparrow$ & \textbf{XCD}$\uparrow$ & \textbf{WCI}$\uparrow$ & \textbf{FaceSim}$\uparrow$ \\
\midrule
Full GroundShot              & \textbf{0.426} & \textbf{0.157} & \textbf{0.264} & \textbf{0.449} \\
\midrule
w/o Entity Grounding         & 0.332 & 0.097 & 0.201 & 0.368 \\
w/o Scheduling               & 0.355 & 0.141 & 0.210 & 0.415 \\
w/o Feedback \& Repair       & 0.387 & 0.143 & 0.214 & 0.421 \\
w/o Anchor Protection        & 0.375 & 0.132 & 0.224 & 0.403 \\
w/o Layered Prompt           & 0.401 & 0.136 & 0.238 & 0.432 \\
w/o VLM Selection            & 0.411 & 0.149 & 0.250 & 0.439 \\
\bottomrule
\end{tabular*}
\end{table}

\paragraph{Effect of entity grounding (w/o Entity Grounding).}
Replacing ReferDINO-based entity grounding with naive first-frame extraction removes entity-specific cropping. The system uses the full first frame as a reference, which contains all entities and background---diluting the identity signal for any single entity. This causes the largest overall degradation: ARC drops by $-$0.094, XCD by $-$0.060, and FaceSim by $-$0.081, confirming that entity-level grounding is the most critical component for both anchor quality and multi-character discrimination.

\paragraph{Effect of scheduling (w/o Scheduling).}
Reverting to narrative-order execution forces the system to bootstrap from whatever reference the first appearance provides, even when that appearance is visually poor (e.g., a wide establishing shot). ARC drops by $-$0.071, the second-largest single-component impact, while WCI drops by $-$0.054, indicating that scheduling specifically prevents worst-case anchor failures. The smaller FaceSim drop ($-$0.034) confirms that scheduling primarily benefits anchor-relative evaluation rather than average pairwise similarity.

\paragraph{Effect of remaining components.}
The remaining four components each provide targeted improvements. Removing feedback and repair primarily hurts worst-case robustness (WCI), as occasional catastrophic generations go uncorrected and enter the reference bank. Disabling anchor protection causes progressive drift---each shot conditions on the previous output rather than the stable early-established identity, accumulating errors over longer sequences. Without layered prompting, naive concatenation of global and shot-level text causes entity leakage, where characters from other shots appear prematurely. Finally, replacing VLM-based reference selection with InsightFace scoring removes context-aware pose matching but has the smallest overall impact, suggesting that entity grounding already provides sufficiently discriminative references for most cases.

\paragraph{Effect of visual memory dynamic maintenance.}
The first reliable reference provides the canonical identity standard, but it is not always sufficient for later shots. We explore whether visual memory should stop after this canonical reference, or maintain a compact set of canonical-consistent supplementary references. The canonical reference remains the identity standard throughout: it is protected from replacement and used to gate later memory admission. However, it need not always be the only image passed to the Ref2V model. When a target shot requires a facial expression or camera view that is poorly represented by the canonical reference, a supplementary reference that has already passed the canonical identity gate can provide a better conditioning match.
This distinction is even more important for scenes. Character identity can often be summarized by a strong face or body prototype, but a scene can extend across views as the camera or characters move through the environment. A single scene reference usually shows only a partial static view of a location, while later shots may revisit the same environment from different camera directions or continue into newly visible regions.
Freezing scene memory at the first valid reference can therefore cause several failure modes: the generator may overfit to a partial background crop, lock later shots to an incompatible camera direction or unnaturally jump back to the initial local background even after previous shots have already moved to another part of the same scene.
Visual memory dynamic maintenance admits additional high-quality scene references of the same environment from later reliable shots. Thus, while dynamic maintenance is useful for character expression and view changes, it is often more critical for scenes because scene consistency usually cannot be reduced to one best canonical image.

\begin{figure}[t]
    \centering
    \includegraphics[width=0.95\linewidth]{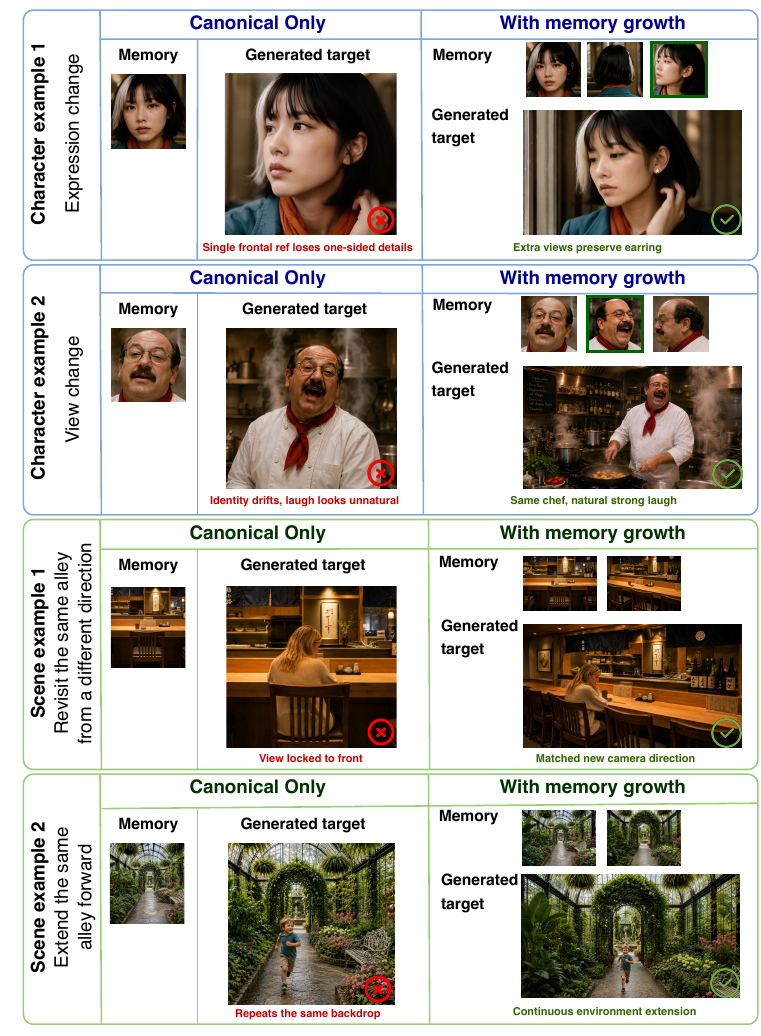}
    \caption{\textbf{Qualitative effect of visual memory dynamic maintenance.}
    The first two rows show character cases where a single canonical reference preserves identity but does not fully cover later expression and view changes; dynamic maintenance provides identity-safe supplementary references for these target shots. The last two rows show scene cases, where dynamic maintenance is more critical: additional scene references help revisit the same place from a different camera direction and extend the environment along the walking path, instead of forcing later shots to reuse a single local background observation.}
    \label{fig:dynamic_maintenance_ablation}
\end{figure}

We therefore qualitatively compare \textbf{Canonical Only}, which freezes entity memory after the first reliable reference, with \textbf{Dynamic Maintenance}, which keeps that canonical anchor while admitting only a few qualified supplementary references. Figure~\ref{fig:dynamic_maintenance_ablation} shows that compact dynamic maintenance remains useful under realistic Ref2V reference limits: these supplementary references provide canonical-consistent coverage for later expressions, views, or scenes, and the scene gains are stronger because multiple references capture complementary regions and camera directions of the same location.

% ============================================================================
\subsection{Human Evaluation}

\begin{table}[t]
\centering
\caption{Human evaluation results. Participants rate each method's multi-shot videos on a 1--5 Likert scale across three dimensions. We report mean scores over 40 scripts; higher is better. GroundShot achieves the highest ratings on all dimensions.}
\label{tab:human_eval}
\vspace{0.3em}
\footnotesize
\setlength{\tabcolsep}{3pt}
\begin{tabular*}{\columnwidth}{@{\extracolsep{\fill}}L{0.40\columnwidth}|ccc}
\toprule
\textbf{Method} & \textbf{Identity}$\uparrow$ & \textbf{Text Align.}$\uparrow$ & \textbf{Overall}$\uparrow$ \\
\midrule
HoloCine              & \underline{3.62} & \underline{3.76} & \underline{3.68} \\
MultiShotMaster-1.3B  & 2.78 & 3.55 & 3.18 \\
MultiShotMaster-14B   & 3.21 & 3.70 & 3.34 \\
StoryMem              & 3.48 & 3.72 & 3.60 \\
\midrule
\textbf{GroundShot (Ours)}  & \textbf{4.18} & \textbf{4.23} & \textbf{4.21} \\
\bottomrule
\end{tabular*}
\end{table}

We conduct a questionnaire-based human evaluation with 48 participants on 40 randomly sampled scripts from GroundBench, covering all four diagnostic modules. For each script, participants are shown anonymized multi-shot videos generated by all methods in randomized order. Each participant independently rates every video on a 1--5 Likert scale along three dimensions: \emph{character identity consistency} (whether recurring characters preserve their appearance across shots), \emph{text alignment} (whether each shot matches its description), and \emph{overall quality} (the perceived quality of the complete multi-shot video).

Table~\ref{tab:human_eval} reports the mean scores. GroundShot achieves the highest ratings across all three dimensions, scoring 4.18 on identity consistency. Notably, GroundShot also leads in text alignment (4.23 vs.\ 3.76), indicating that entity-grounded generation does not sacrifice prompt adherence for consistency. The overall quality scores closely track identity ratings, suggesting that human perception of multi-shot video quality is strongly driven by character consistency.

% ============================================================================
\subsection{Discussion}

\begin{figure}[t]
    \centering
    \includegraphics[width=0.98\linewidth]{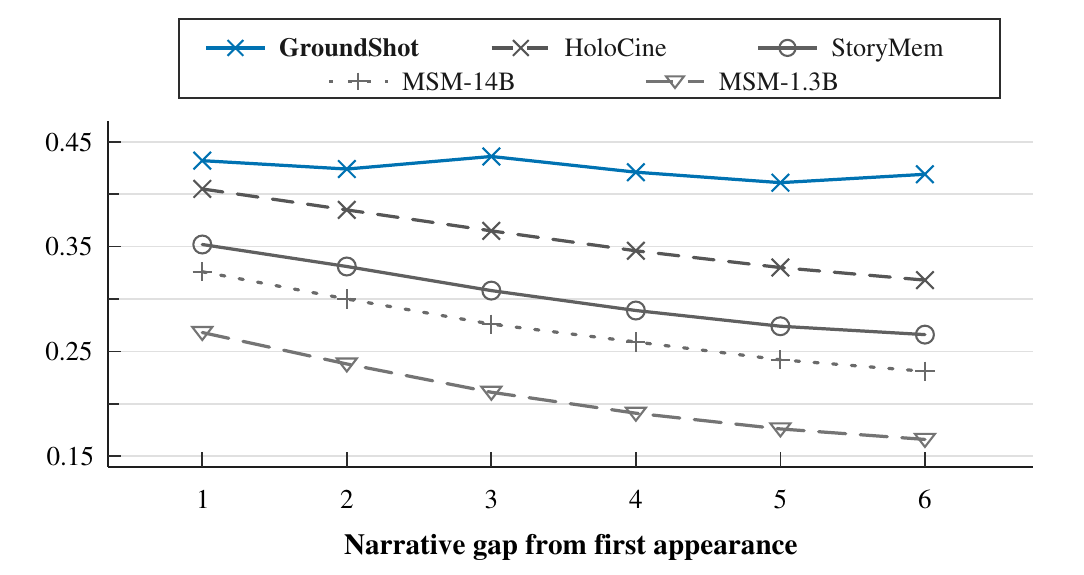}
    \caption{Anchor-relative consistency over narrative distance. The x-axis reports the narrative shot gap from the first valid appearance, and the y-axis reports ARC (higher is better). Baselines tend to decay as the gap increases, whereas GroundShot remains stable with small non-monotonic variation because generation order is decoupled from narrative order and later shots can condition on high-quality canonical references.}
    \label{fig:arc_curve}
\end{figure}

\paragraph{Consistency over narrative distance.}
Figure~\ref{fig:arc_curve} visualizes how ARC changes as recurring entities move farther from their first valid appearance in the final narrative order. For methods that follow narrative order or rely on frame/keyframe memory, ARC tends to decrease with larger gaps, reflecting memory aging and accumulated identity drift. GroundShot shows a different pattern: its ARC stays in a narrow high range rather than forming a monotonic decay. This is expected because the scheduler may generate an informative close-up or medium shot before its narrative neighbors, establish it as a canonical reference, and then condition later narrative positions on that same reference. The remaining variation mainly reflects local shot difficulty, such as occlusion, scale, viewpoint, or interaction, rather than distance along a generation chain.

\paragraph{Reliability of scene references.}
Location memory has a different failure mode from character and object memory. A scene reference cannot be obtained by simply cropping a foreground entity; it must remove foreground subjects and preserve or reconstruct the background. If this reconstruction leaves blocky regions, smeared textures, or residual human figures, reference-conditioned generation may reproduce these artifacts in later shots. GroundShot therefore admits a reconstructed scene only when validation confirms that the unmasked background remains intact, the masked region is plausibly filled, and no prominent foreground subject remains. When this validation fails, especially in close-up shots with little usable background, the system skips the location update rather than storing a weak scene reference. This is usually preferable because such close-up backgrounds carry limited scene identity information, while text prompts can still preserve lighting and setting context until a clearer scene view appears.

\paragraph{Computational overhead and trade-offs.}
GroundShot introduces additional test-time computation because generated shots may be verified by a VLM critic and regenerated when severe failures are detected. This overhead is bounded and measured explicitly. On GroundBench, GroundShot averages 1.20 video generation calls per final shot, with 83.1\% of shots passing on the first attempt; regeneration is therefore occasional rather than the default.
With batched script parsing and risk-gated verification, GroundShot uses 2.3 VLM critique calls and 3.4 text-only LLM calls per script on average, amortized across roughly six shots. Entity-level grounding runs locally with ReferDINO.

To verify that the gains do not simply come from extra generations, we evaluate GroundShot without feedback and repair (Table~\ref{tab:ablation}). This one-generation-per-shot variant still outperforms the strongest external baselines on ARC and FaceSim, showing that scheduling and grounded memory provide the main consistency gains; the full critique-and-retry loop mainly improves worst-case robustness. When latency or budget is the primary constraint, users can disable this loop and retain most of the consistency benefit. GroundShot still inherits the quality ceiling of its video generation backend: if the backend cannot produce a recognizable entity in any single attempt, retries may not resolve the issue.

% ============================================================================
\section{Conclusion}

We presented GroundShot, a training-free and model-agnostic agentic framework for visually consistent multi-shot video generation. GroundShot builds entity-level visual memory from its own generated outputs, using quality-aware scheduling, grounding, verification, canonical-reference protection, and reference selection to decouple production order from narrative order and reduce cross-shot drift. We also introduced GroundBench, an entity-grounded diagnostic benchmark for fine-grained evaluation of multi-shot consistency. Across automatic metrics, same-backend comparisons, ablations, and human evaluation, GroundShot improves entity-level consistency without additional training or model modification, with entity grounding and scheduling contributing the largest gains.

GroundShot remains limited by the quality ceiling of its underlying generator. Future work can extend this reference-management perspective to longer narratives, richer scene memory, and joint reasoning over consistency, temporal coherence, and efficiency.

\clearpage
\setcounter{page}{1}
\maketitlesupplementary

\section{Supplementary Method Details}

\subsection{Entity Graph Construction}
\label{sec:supp_entity_graph}

GroundShot relies on a lightweight entity graph to connect script-level semantics with visual grounding and reference reuse. For each shot $s_i$, the LLM parser outputs a set of entity records $\mathcal{E}_i$. Each record contains:
\begin{itemize}
\raggedright
    \item a stable entity ID (e.g., \texttt{char\_alex}, \texttt{obj\_briefcase}, or \texttt{loc\_station});
    \item an entity type: \texttt{character}, \texttt{object}, or \texttt{location};
    \item a natural-language description used for grounding and prompt construction;
    \item a grounding priority (high for recurring characters and important objects; medium for locations);
    \item aliases used for cross-shot coreference.
\end{itemize}

\paragraph{Cross-shot coreference.}
The parser receives the known entities discovered in previous shots. This allows different textual mentions to map to the same entity ID. For example, consider the following script:
\begin{quote}
Shot 1: ``Alex Chen, a male detective in a dark trench coat, enters the police station lobby.''\\
Shot 2: ``The detective sits across from a nervous suspect.''\\
Shot 3: ``He walks into a rain-soaked alley at night.''
\end{quote}
The parser assigns the three mentions ``Alex Chen'', ``the detective'', and ``he'' to the same entity \texttt{char\_alex}. A simplified output is:
\begingroup
\footnotesize
\begin{verbatim}
S1:
  char_alex:
    type: character
    desc: "Alex Chen, detective"
  loc_station:
    type: location
    desc: "police lobby"

S2:
  char_alex:
    type: character
    aliases: ["detective"]
  char_suspect:
    type: character
    desc: "nervous suspect"
  loc_room:
    type: location
    desc: "interrogation room"

S3:
  char_alex:
    type: character
    aliases: ["he"]
  loc_alley:
    type: location
    desc: "rain-soaked alley at night"
\end{verbatim}
\endgroup

\paragraph{Entity counts.}
The parser also extracts expected entity counts when they are explicit or strongly implied by the script. These counts are used by the runtime verification module. For example, ``three guards stand at the gate'' yields an expected character count of three; if the generated video contains a different number of people, the system can repair the shot before grounding the result into the memory.

\paragraph{Global context filtering.}
The optional global caption is not directly appended to every shot prompt. Instead, GroundShot extracts only global visual context such as visual style, mood, setting, and narrative atmosphere. Concrete entity descriptions are handled through the per-shot entity graph. This prevents future entities in the story summary from leaking into earlier prompts. For example, if a masked rider first appears in shot 5, the prompt for an indoor dialogue in shot 2 should not mention the rider; otherwise the generator may introduce that character before the script does. By separating global context from shot-specific entities, GroundShot keeps style consistent while preserving shot-level content control.

\paragraph{Shot-entity incidence graph.}
After parsing all shots, GroundShot obtains an incidence matrix $M_{i,e}$ indicating whether entity $e$ appears in shot $s_i$. This graph is used by the scheduler to predict which shots can provide references for which entities, and by GroundShot to query the entity-level visual memory for the references needed by the current shot.

% ============================================================================
\section{GroundBench: Detailed Specification}
\label{sec:supp_groundbench}

\subsection{Benchmark Taxonomy and Distribution}
\label{sec:supp_groundbench_taxonomy}

GroundBench contains 54 YAML scripts and 309 shots.  It is organized into 18 sub-modules under four diagnostic modules; each sub-module has three scripts with \texttt{challenge\_level}$\in\{1,2,3\}$.

\begin{table*}[t]
\centering
\caption{\textbf{GroundBench taxonomy.}  ``Chars.'' and ``Objs.'' count script-level tracked foreground definitions.}
\label{tab:supp_groundbench_modules}
\footnotesize
\setlength{\tabcolsep}{4pt}
\begin{tabular*}{\textwidth}{@{\extracolsep{\fill}}L{0.08\textwidth}L{0.22\textwidth}L{0.36\textwidth}ccc}
\toprule
\textbf{Module} & \textbf{Diagnostic Axis} & \textbf{Sub-modules} & \textbf{Scripts} & \textbf{Shots} & \textbf{Chars./Objs.} \\
\midrule
A & Identity preservation under visual transformations
& A1 scale, A2 lighting, A3 angle, A4 temporal gap, A5 occlusion
& 15 & 81 & 15 / 0 \\
B & Multi-identity discrimination
& B1 distinct pairs, B2 similar pairs, B3 groups, B4 character--object consistency, B5 swap stress
& 15 & 81 & 39 / 6 \\
C & Structural and editing patterns
& C1 scene revisit, C2 parallel timelines, C3 shot-type diversity, C4 long sequence, C5 late introduction
& 15 & 102 & 24 / 0 \\
D & Robustness probes
& D1 style transition, D2 crowd disambiguation, D3 pronoun coreference
& 9 & 45 & 9 / 0 \\
\bottomrule
\end{tabular*}
\end{table*}

The level split is balanced at the script level (18 scripts per level), with 99/103/107 shots for levels 1/2/3.  The foreground metadata contains 87 character definitions and 6 recurring object definitions; recurring objects appear only in B4.

\subsection{Unified Evaluation Detector (UED) Configuration}
\label{sec:supp_ued}

UED is method-independent: it reads only the released script and generated videos, never method-specific references, registries, or debug crops.
\begin{itemize}
    \item Sample 8 uniformly spaced frames per shot.
    \item Run ReferDINO~\cite{liang2025referdino} for each expected foreground entity using the YAML entity name and description.
    \item Accept a crop only when grounding confidence and query-token alignment pass fixed thresholds; otherwise the expected shot-entity pair is marked undetected.
    \item Use ArcFace~\cite{insightface} for valid face crops and DINOv2~\cite{oquab2023dinov2} for non-face character crops and objects.
    \item Background features: ESC uses the same sampled frames and UED foreground boxes to mask characters and objects before extracting DINOv2 background embeddings.
\end{itemize}

\subsection{Metric Definitions}
\label{sec:supp_metrics}

GroundBench reports ARC, WCI, XCD, EAF, and ESC as entity-grounded metrics, plus FaceSim, CSC, and VBench aesthetic/dynamic scores as reference metrics.  Metrics are computed from UED outputs; entities, settings, or pairs without valid evidence are excluded from the corresponding metric.

\paragraph{Evidence and calibration.}
Each accepted crop has a DINOv2 embedding and, when a valid face is detected, an ArcFace embedding.  ArcFace and DINOv2 cosine scores are separately calibrated with held-out data to a shared same-entity score in $[0,1]$.  For same-entity shot pairs, ARC/WCI use ArcFace when both crops contain valid faces and DINOv2 otherwise, then take the median calibrated crop-pair score across sampled frames.

\paragraph{ARC and WCI.}
\textbf{ARC} averages each recurring entity's similarity to its first valid detected appearance.  \textbf{WCI} uses the same anchor-relative comparisons but reports the weakest valid match, making it sensitive to catastrophic drift.

\paragraph{XCD.}
\textbf{XCD} is the margin between same-character consistency and different-character confusion.  Different-character similarities are averaged per unordered character pair before aggregation, so densely detected pairs do not dominate.  We also report \texttt{xcd\_face} and \texttt{xcd\_dino}.

\paragraph{EAF.}
\textbf{EAF} scores CLIP similarity between each accepted crop and the entity's textual appearance description, rescales it to $[0,1]$, averages within each entity, and then averages across valid entities.

\paragraph{ESC.}
\textbf{ESC} masks foreground entities, extracts DINOv2 background features, and averages pairwise similarities within each recurring setting.  \texttt{esc\_all} applies the same comparison without setting groups.

\paragraph{Reference metrics.}
FaceSim averages same-character ArcFace similarities over valid face pairs.  CSC averages pairwise ViCLIP shot-video similarities and is therefore holistic rather than entity-grounded.  VBench aesthetic and dynamic-degree scores are averaged over shots to monitor visual quality and motion strength.

% ============================================================================
\section{Entity-Level Visual Memory}
\label{sec:supp_visual_memory}

The entity-level visual memory $\mathcal{R}=\{\mathcal{R}_e\}_e$ stores a compact active reference set for each entity.  Characters and objects use foreground crops; locations use reconstructed scene references.  Each entity has one protected canonical reference $r^*_e$ and, after canonical initialization, a small auxiliary pool for canonical-consistent view, lighting, pose, or scene-coverage variation.  The first active reference written for an entity is always its canonical reference.

\subsection{Multi-Gate Registration}
\label{sec:supp_registration}

When visual grounding produces a foreground crop $c$ for entity $e$, GroundShot first applies type-specific validation.  The candidate must satisfy a composite quality threshold $q(c)\geq\tau_{\min}=0.4$, computed from sharpness, occlusion, and type-specific visibility terms.  Character crops additionally require face detection confidence $\geq0.3$; object crops are checked by a VLM for recognizability and agreement with the entity description.

If $e$ has no canonical reference, $c$ enters memory only when it satisfies the stricter canonical-quality conditions in Sec.~\ref{sec:supp_canonical_reference}.  Otherwise it is rejected.  If $r^*_e$ already exists, $c$ must remain consistent with it:
\begin{equation}
    \operatorname{sim}(c,r^*_e) \geq \theta_{\mathrm{id}} .
\end{equation}
Finally, the candidate must add non-redundant evidence.  Using CLIP ViT-B/32 embeddings $\mathbf{e}$, GroundShot rejects $c$ if
$\max_{r\in\mathcal{R}_e}\cos(\mathbf{e}_c,\mathbf{e}_r)\geq0.92$.  Each shot may contribute at most two references for the same entity, and each entity keeps at most six active references: one canonical reference and up to five auxiliaries.

\subsection{Location Reference Admission}
\label{sec:supp_location_reference}

For a shot containing a location entity, GroundShot masks grounded foreground characters and objects, dilates the union mask, and uses an instruction-following image editor to reconstruct the occluded background.  The edited frame becomes a candidate scene reference $c_{\mathrm{loc}}$.  It is admitted only if it has sufficient usable background, matches the parsed location description, and contains no prominent foreground remnants or reconstruction artifacts.  This produces a location quality score $q_{\mathrm{loc}}(c_{\mathrm{loc}},e)$; canonical consistency is then checked with scene-level similarity and coverage.  Failed validation leaves the location memory unchanged.

\subsection{Canonical Reference Initialization}
\label{sec:supp_canonical_reference}

For each entity, the canonical slot remains empty until a candidate satisfies all canonical-quality conditions:
\begin{enumerate}
    \item High quality: $q(c) \geq 0.85$ for foreground references, or $q_{\mathrm{loc}}(c_{\mathrm{loc}},e) \geq 0.85$ for location scene references;
    \item Canonical-ready visibility: face confidence $\geq 0.7$ and near-frontal visibility for characters; recognizable, complete appearance for objects; artifact-free, foreground-free reconstruction for locations;
    \item Empty canonical slot: no canonical reference already exists for this entity.
\end{enumerate}
The first candidate satisfying these conditions initializes $r^*_e$, regardless of narrative or execution index.  Before $r^*_e$ exists, non-canonical candidates are rejected rather than stored as auxiliaries.  After $r^*_e$ exists, later candidates may enter only as auxiliary references and cannot overwrite it.

\subsection{Eviction Mechanisms}
\label{sec:supp_eviction}

Eviction applies only to auxiliary references.  When the pool is full, GroundShot first removes auxiliaries that are low-quality ($q<0.5$) or redundant with a retained reference.  If the pool remains full, a new reference replaces the lowest-quality auxiliary only when its quality score is higher; otherwise it is rejected.  The same cleanup runs every 5 executed shots.

\subsection{Query Strategy}
\label{sec:supp_query}

For a target shot, visual memory returns the top-$k$ active entries sorted by canonical status first, then context-aware utility when available (viewpoint, lighting, pose, location coverage, and shot type), otherwise quality score, with shot ID used only as a final tie-breaker.  This keeps the canonical identity standard first while still allowing useful auxiliary variation.

% ============================================================================
\section{Agentic Reference Selection}
\label{sec:supp_agentic_ref_selection}

When multiple references are available, GroundShot selects a target-aware subset rather than blindly using the highest-quality image.  The canonical reference remains the default consistency anchor, but it is not mandatory in every generation call: characters usually keep it for identity, while objects and locations may use an auxiliary-only or multi-auxiliary subset when another view better matches the target shot.

\subsection{Motivation and Modes}
\label{sec:supp_ref_sel_motivation}
\label{sec:supp_ref_sel_modes}

The deterministic baseline ranks candidates with type-specific low-level scores.  For characters, it uses
\begin{equation}
    q_{\text{trad}}(c)=0.4\,\sigma(c)+0.4\,d(c)+0.2\,f(c),
    \label{eq:trad_score}
\end{equation}
where $\sigma$ is normalized sharpness, $d$ is face-detection confidence, and $f=\max(0,1-|\mathrm{yaw}|/90^\circ)$ is frontality.  Object and location scores replace the face terms with recognizability, completeness, coverage, and artifact penalties.  These scores are useful for filtering poor references, but they do not capture whether a reference matches the target action, viewpoint, lighting, or scene extent.

GroundShot therefore supports three modes.  \emph{Traditional} mode uses Eq.~\ref{eq:trad_score} and deterministic type-aware rules.  \emph{Agent} mode asks a VLM to choose the subset from the canonical reference and auxiliary candidates.  \emph{Hybrid} mode, used by default, runs the agent selector when available and falls back to the traditional scorer on timeout or API failure.

\subsection{Selector Interface}
\label{sec:supp_ref_sel_prompt}
\label{sec:supp_ref_sel_parsing}
\label{sec:supp_ref_sel_integration}

For each entity in the current shot, GroundShot retrieves the canonical reference, if initialized, and up to $K_{\max}-1$ auxiliary candidates using the query ordering in Sec.~\ref{sec:supp_query}.  The agent receives these images, the entity description, entity type, shot type, and shot text.  The prompt asks it to preserve identity or scene consistency while selecting references that best match the requested view, action, lighting, or location coverage.

The VLM returns a compact JSON decision:
\begin{verbatim}
{
  "use_canonical": <bool>,
  "selected_indices": [<int, ...>],
  "selection_mode": "<canonical_only | canonical_plus_aux | aux_only | multi_aux | none>",
  "confidence": <float in [0,1]>,
  "reason": "<string>",
  "analysis": {
    "candidate_1": "<string>",
    ...
  },
  "alternatives": [<int>, ...]
}
\end{verbatim}
\texttt{selected\_indices} is 0-based over auxiliary candidates.  Empty \texttt{selected\_indices} with \texttt{use\_canonical=true} means canonical-only; empty \texttt{selected\_indices} with \texttt{use\_canonical=false} means no reference is selected for that entity.  Invalid JSON, out-of-range indices, or low-confidence outputs trigger graceful degradation to heuristic extraction and then to the traditional selector.  The final per-entity subsets are packed under the backend's global reference limit, prioritizing character identity anchors while allowing object or location auxiliaries to replace a canonical reference when they are more useful for the target shot.

\end{document}